\pdfoutput=1

\documentclass[11pt]{article}

\usepackage[]{emnlp2021}
\usepackage{multirow}

\usepackage{times}
\usepackage{latexsym}
\usepackage{multirow}
\usepackage{graphicx}

\usepackage[T1]{fontenc}
\usepackage[utf8]{inputenc}

\usepackage{microtype}

\usepackage{csquotes}

\definecolor{ForestGreen}{RGB}{34,139,34}

\title{Implicit Premise Generation with Discourse-aware Commonsense Knowledge Models}

\author{Tuhin Chakrabarty \textsuperscript{1},
  Aadit Trivedi\textsuperscript{2}, 
  \textbf{and} \textbf{Smaranda Muresan}\textsuperscript{1,3}\\ 
  \textsuperscript{1}Department of Computer Science, Columbia University \\
  \textsuperscript{2}College of Computing, Georgia Tech\\
  \textsuperscript{3}Data Science Institute, Columbia University\\
  {\tt atrivedi@gatech.edu}, {\tt \{tuhin.chakr,smara\}@cs.columbia.edu}
  }

\begin{document}
\maketitle
\begin{abstract}
Enthymemes are defined as arguments where a premise or conclusion is left implicit. We tackle the task of \emph{generating the implicit premise in an enthymeme}, which requires not only an understanding of the stated conclusion and premise, but also additional inferences that could depend on commonsense knowledge. The largest available dataset for enthymemes \cite{habernal-etal-2018-argument} consists of 1.7k samples, which is not large enough to train a neural text generation model. To address this issue, we take advantage of a similar task and dataset: Abductive reasoning in narrative text \cite{Bhagavatula2020Abductive}. However, we show that simply using a state-of-the-art seq2seq model fine-tuned on this data might not generate meaningful implicit premises associated with the given enthymemes.  We demonstrate that encoding discourse-aware commonsense during fine-tuning improves the quality of the generated implicit premises and outperforms all other baselines both in automatic and human evaluations on three different datasets.
\end{abstract}

\section{Introduction}

In argumentation theory, an enthymeme is defined as an incomplete argument found in discourse, where some components are explicit, but other propositions are left implicit and need to be \emph{filled in} as premises or conclusions to fully understand what the argument is \cite{waltonReed2005}. In many instances the missing proposition is a premise. 
The well-cited example of the Silver Blade case from one of Sherlock Holmes' stories \cite{waltonReed2005} presents such as an incomplete argument 
\begin{displayquote}
A dog was kept in the stable, and yet, though someone had been in and fetched out a horse, he had not barked enough to rouse the two lads in the loft. Obviously, the midnight visitor was someone whom the dog knew well.
\end{displayquote}

The missing premise in this case is the generalization ``Dogs generally bark when a person enters an area unless the dog knows the person well." 

While there has been work on identification (i.e., classification) and reconstruction of implicit premises in enthymemes \cite{rajendran-etal-2016-contextual,habernal-etal-2018-argument, reisert-etal-2015-computational,boltuzic-snajder-2016-fill,Razuvayevskaya2017FindingEI}, to our knowledge, \emph{automatically generating an implicit premise from a given enthymeme} is a new task. 
There are two main challenges that need to be addressed: 1) lack of large scale data of incomplete arguments together with annotated missing premises needed to train a sequence-to-sequence model (the largest such set contains 1.7K instances \cite{habernal-etal-2018-argument}); and 2) the inherent need to model commonsense or word knowledge. 

We propose an approach for generating an implicit premise given a incomplete argument that aims to address these two challenges.  Our contributions are three fold.

\begin{table}[]
\small
\centering
\begin{tabular}{|l|l|l|}
\hline
Reason                                                                                 & \multicolumn{2}{l|}{Vaccinations save lives}                                                                                \\ \hline
Claim                                                                                  & \multicolumn{2}{l|}{\begin{tabular}[c]{@{}l@{}}Vaccination should be mandatory\\ for all children\end{tabular}}             \\ \hline\hline
ZeroShot                                                                               & \multicolumn{2}{l|}{Vaccines save lives, they save money}                                                                   \\ \hline
\begin{tabular}[c]{@{}l@{}}Fine-tuned on \\\textit{ART} \end{tabular}                 & \multicolumn{2}{l|}{\begin{tabular}[c]{@{}l@{}}Vaccinations are the best way to \\ protect children.\end{tabular}}          \\ \hline
\begin{tabular}[c]{@{}l@{}}Fine-tuned on \\ \textit{ART +PARA-C}\end{tabular} & \multicolumn{2}{l|}{\begin{tabular}[c]{@{}l@{}}Vaccinations are the best way to\\ prevent childhood diseases.\end{tabular}} \\ \hline
\end{tabular}
\caption{\label{table:1}Implicit Premise Generation by BART \cite{lewis2019bart} in three different setting for an input enthymeme from dataset by \citet{habernal-etal-2018-argument} }
\vspace{-3ex}
\end{table}

\emph{A new task of generating an implicit premise given an incomplete argument (enthymeme).} Given an enthymeme consisting of a stated conclusion and a stated premise, generate the implicit/missing premise.  As the backbone sequence-to-sequence architecture we use BART \cite{lewis2019bart}. 

\emph{Leverage abductive reasoning as an auxiliary task}. To address the first challenge, we rely on an observation from argumentation theory that incomplete arguments in naturally occurring discourse, more often than not, require abductive reasoning (plausible explanations) rather than the more strict form of reasoning based on deductive logic \cite{waltonReed2005,10.2307/40320292}. The Silver Blaze case is such an example. 
We leverage the Abductive Reasoning in Narrative Text (\textit{ART}) dataset introduced by \citet{Bhagavatula2020Abductive} to fine-tune a BART model. \textit{ART} consists of pairs of observations together with the plausible explanation to be generated (Section \ref{section:data}).    

\emph{Encoding discourse-aware common sense knowledge.} To address the second challenge, we rely on PARA-COMET \cite{Gabriel2021ParagraphLevelCT}, a discourse-aware knowledge model that incorporates paragraph-level information to generate coherent commonsense inferences from narratives. We encode the outputs of PARA-COMET during fine-tuning BART on our auxillary dataset (ART) (Section \ref{section:method}). We show on three different datasets (Section \ref{section:data}) that this knowledge-enhanced model performs best both in automatic and human-based evaluations (Section \ref{section:eval}).   

Table \ref{table:1} shows an example of an enthymeme consisting of a stated premise and conclusion and the generated implicit premise by a BART model (zero-shot), by a BART model fine-tuned on \textit{ART} dataset, and a BART model fine-tuned on \textit{ART} augmented with discourse-aware commonsense knowledge derived from PARA-COMET. We make the code available at \url{https://github.com/tuhinjubcse/EnthymemesEMNLP2021}.

\section{Related Work}

Prior work on enthymeme reconstruction has focused primarily on the identification (i.e., classification) of implicit premises in enthymemes \cite{rajendran-etal-2016-contextual,habernal-etal-2018-argument, reisert-etal-2015-computational,boltuzic-snajder-2016-fill,Razuvayevskaya2017FindingEI}.  \citet{boltuzic-snajder-2016-fill} study how to identify enthymemes in online discussions, while  \citet{habernal-etal-2018-argument} present the task of identifying the correct warrant given two candidates warrants in order to reconstruct an enthymeme.  \citet{rajendran-etal-2016-contextual} introduce an approach to classify the stance of a statement as implicit or explicit, as a first step towards the long term goal of enthymeme reconstruction. 
Unlike these works which propose discriminative approaches to identify an enthymeme or the (correct) implicit premises, we focus on generative models that aim to \emph{generate an implicit premise given an enthymeme}, using abductive reasoning and discourse-aware commonsense knowledge.

\citet{alshomary-etal-2020-target} introduce a closely related task of generating an argument's conclusion from its premises. Specifically, they focus on the subtask of inferring the conclusion’s target from the premises. They develop two complementary target inference approaches: one ranks premise targets and selects the top-ranked target as the conclusion target, the other finds a new conclusion target in a learned embedding space using a triplet neural network. Unlike this paper, our work focuses on a new task of generating an implicit premise given an enthymeme that consists of a stated conclusion and a stated premise.

\label{section:data}
\section{Datasets} \label{section:data}

\paragraph{Training dataset.} 
Based on the theoretical connection between enthymemes and abductive reasoning, we use the \textit{Abductive Reasoning in narrative Text (ART)} data developed for the abductive NLG task \cite{Bhagavatula2020Abductive} 
to train our models. The task is framed as: given two observations (O1 and O2) from a narrative,  generate the most  plausible explanation (hypothesis) (Table \ref{table:anlg}).  The observations O1, O2 in \textit{ART} are drawn from the ROCStories \cite{mostafazadeh2016corpus} dataset, a large collection of short, manually curated five sentence stories. 
The beginning and ending of each story maps to the first (O1) and second (O2) observations in ART, respectively. \citet{Bhagavatula2020Abductive} presented O1 and O2 as narrative context to crowdworkers and prompted them to generate plausible and implausible Hypotheses (H) to explain the observations. To avoid  annotation artifacts, \citet{Bhagavatula2020Abductive}  applied an adversarial filtering step to retain one challenging pair of plausible and implausible hypotheses that are hard to distinguish between. 
The \textit{ART} training set consists of 50481 instances, while the validation and test set consist of 7252 and 14313 instances,  respectively.
As can be seen in Table \ref{table:anlg} the observations O1 and O2 could be "mapped" to the stated Premise and the stated Claim in an enthymeme, while the hypothesis H is mapped to the \emph{implicit premise} we try to generate.

\begin{table}[t]
\small
\centering
\begin{tabular}{|l|l|l|}
\hline

O1 & \multicolumn{2}{l|}{Alex had his heart set on an ivy league college}                                                          \\ \hline
O2 & \multicolumn{2}{l|}{\begin{tabular}[c]{@{}l@{}}Alex ended up achieving his dream \\ of getting into the school.\end{tabular}} \\ \hline
H  & \multicolumn{2}{l|}{Alex applied to Harvard}                                                                                  \\ \hline
\end{tabular}
\caption{\label{table:anlg}Instances from the 
\textit{ART} dataset.} 
\vspace{-3ex}

\end{table}

\paragraph{Test datasets.}
We test our models on three different datasets of incomplete arguments (enthymeme) annotated with human-generated implicit/missing premises. 
First, we use the Argument Reasoning Comprehension Task dataset released by \citet{habernal-etal-2018-argument} (\textbf{D1}), which contains 1654 \textit{\{claim, premise, warrant(implicit premise)\}} triples. Second, we used the dataset introduced by \citet{boltuzic-snajder-2016-fill}, which contains 494 enthymemes from an online debate forum with human annotated implicit premises (\textbf{D2}).  Third, we use the dataset introduced by \citet{becker-etal-2020-implicit} (\textbf{D3}), which contains implicit premises annotated for each  arguments from the MicroText Corpus \cite{peldszus2015annotated}. For \textbf{D3}, we focus only arguments that are in a \emph{support} relation since this corresponds to our task. Moreover, we choose the cases where there is only one implicit premise, rather than a chain of linked premises. This results in a total of 112 enthymemes for D3. 
For all datasets, we apply automatic filtering to keep only full-formed sentences as claim and premises (e.g., remove cases where the stated premise/claim consists of a noun-phrase, a partial clauses, or many sentences).

\begin{table}[]
\centering
\small
\begin{tabular}{|l|l|}
\hline
\begin{tabular}[c]{@{}l@{}}Encoder\\ Input\end{tabular}               & \begin{tabular}[c]{@{}l@{}}Amy was looking through her \\ mother's old scrapbooks. \ \textbf{[SEP]} Amy \\ realized her mother had dated her\\  history professor.\end{tabular}                                                                           \\ \hline\hline
\begin{tabular}[c]{@{}l@{}}Encoder\\ Input +\\ PARA-COMET\end{tabular} & \begin{tabular}[c]{@{}l@{}}Amy was looking through her \\ mother's old scrapbooks. \ \textbf{[SEP]} \textbf{\color{ForestGreen}to} \\ \textbf{\color{ForestGreen}find something} \ \textbf{[SEP]} Amy realized\\ her mother had dated her history\\ professor.\end{tabular}                                                      \\ \hline\hline
\begin{tabular}[c]{@{}l@{}}Decoder\\ Ouput\end{tabular}               & \begin{tabular}[c]{@{}l@{}}Amy was looking through her\\ mother's old scrapbooks. \textbf{\textit{\color{blue}And since}} \\ \textit{Amy found pictures of her history}\\ \textit{professor and mother together}. Amy\\ realized her mother had dated her\\ history professor.\end{tabular} \\ \hline
\end{tabular}
\caption{\label{table:finetuning} Encoder input in two settings: fine-tuning on \textit{ART} 
and fine-tuning on \textit{ART} + PARA-COMET (the green text between [SEP]).
For decoder's output every hypothesis is prepended by \textit{And since} in bolded blue.}
\vspace{-3ex}
\end{table}

\section{Method}\label{section:method}

For our generation model, we use BART \cite{lewis2019bart}, a pre-trained conditional language model that combines bidirectional and auto-regressive transformers. It is implemented as a sequence-to-sequence model with a bidirectional encoder over corrupted text and a left-to-right auto-regressive decoder.  

\paragraph{Fine-tuning BART on \textit{ART}.} To fine-tune BART on the \textit{ART} dataset (Section \ref{section:data}), we concatenate O1 and O2 with a special delimiter [SEP] as input to BART encoder as shown in Table \ref{table:finetuning} Row 1. For decoding, we focus on reconstructing the entire argument given an enthymeme.
To encourage fluency and coherence in our generated argument, we prepend the plausible hypothesis (implicit premise) with a discourse marker \textit{And since} (Table 3 Row 3) during fine-tuning.

\paragraph{Fine-tuning BART on PARA-COMET enhanced \textit{ART}.}
Adapted knowledge models such as COMET \cite{bosselut-etal-2019-comet} have been shown to generate implicit commonsense inferences along several dimensions (depending on what knowledge graphs they were pre-trained on). 
PARA-COMET \cite{Gabriel2021ParagraphLevelCT}, is an extension of COMET pre-trained on ATOMIC  \cite{atomic} that is able to generate discourse-aware common sense knowledge. ATOMIC is a knowledge graph that contains 9 relations related to social commonsense knowledge, including dynamic aspects of events such as causes and effects, if-then conditional statements, and mental states.
Given a text
with T sentences ${S_{1}, S_{2}...S_{T} }$, PARA-COMET generates a set of commonsense inferences for the 9 inferential relations from ATOMIC for each sentence $S_{i}$, which are consistent with the entire narrative.  
Following PARA-COMET's input format, we create a discourse of two sentences containing [O1,O2] from ART. We then feed this as an input to the trained PARA-COMET model and obtain 9 commonsense relations  for both O1 and O2. Given the causal nature of the implicit premises for this work we use only the relation \textit{xIntent}. Given an event (e.g., ``X compliments Y"), \textit{xIntent} states the likely intents of person X (e.g., ``X wants to be nice"). We only consider \textit{xIntent} returned for O1 (Premise on our task). We experimented with other relations as well as \textit{xIntent} for both O1 and O2 but the results were not better.  After obtaining discourse-aware commonsense, we concatenate \{O1, commonsense, O2\} in a sequential order as shown in Table \ref{table:finetuning} Row 2 and pass it to BART's encoder  for fine-tuning. For decoding, we use the same process as before (Table \ref{table:finetuning} Row 3).

\paragraph{Inference-time decoding.} For generation on our task and test sets, we concatenate the \{Premise, Claim\} or \{Premise, commonsense, Claim\} in a given enthymeme in the same way as shown in Table \ref{table:finetuning} and pass as an input to the encoder of fine-tuned BART. The fine-tuned BART model then generates the entire argument along with the implicit premise auto-regressively. We use beam search with a beam width of 5 for generation. Post decoding, we split the argument into 3 individual sentences and treat the middle sentence starting with \textit{And since} as the implicit premise after removing the artificially added discourse marker.

For zero-shot setting,  we use the pre-trained BART (bart-large) model. We use the format \{\emph{Premise. And since [MASK]. Claim}\} and let the language model generate an implicit premise.

\section{Evaluation and Results} \label{section:eval}
We evaluate three setups: 1) directly use pre-trained BART  (Zero-shot); 2) fine-tune BART on \textit{ART}; 3) fine-tune BART on \textit{ART+PARA-COMET}. 

\paragraph{Automatic Evaluation Setup.} We use \textit{BLEU}~\cite{BLEU}, one of the most widely used automatic metrics for generation tasks to compute BLEU-1 and BLEU-2 scores between the system output and the human written gold implicit premise. We also report F1-Score of \textit{BERTScore}, a metric for evaluating text generation using contextualized embeddings. 
\paragraph{Human evaluation setup.} We select 50 enthymemes from each test set (total of 150 enthymemes) and the output of our fine-tune BART models (with or without PARA-COMET). We hired crowdworkers on the Amazon Mechanical Turk platform. Given an enthymemes they were asked if the generated implicit premises were plausible or not (agreement: 0.56 based on Krippendorff's $\alpha$). Each enthymeme was judged for plausibility by 3 distinct Turkers (50 crowdworkers overall). As it was a binary judgement, we took majority voting which means if 2/3 of the annotators thought it was plausible we marked it as plausible.  Plausibility judgement considers whether the generated premise  was grammatical, relevant to the argument, coherent with our commonsense and completes the argument. 

\paragraph{Results.}
While pre-trained language models often contain structured commonsense \cite{davison-etal-2019-commonsense,Zhou2020EvaluatingCI} Table \ref{table:auto} shows that pre-trained BART cannot generate plausible implicit premises. Fine-tuning on the \textit{ART} dataset improves the results significantly. Finally, the model that encodes discourse-aware commonsense outperform all baselines on all test datasets (D1, D2 and D3). Human evaluation
further demonstrates that encoding commonsense knowledge leads to better implicit premise generation (Table \ref{table:human}). 

\paragraph{Analysis.}
We notice that adding commonsense beams from PARA-COMET makes the generated implicit premise more plausible. For instance, for the stated claim and premise from D3 in Table \ref{analysis}, we see that PARA-COMET adds a beam \textit{to feel better}.
Similarly it adds a beam \textit{to learn more} for the stated claim and premise from D1 for both examples shown in Table \ref{analysis}. We posit that adding these in combination with the stated claim and premise, leads our model to infer more plausible implicit premises compared to the ones generated by BART fine-tuned on \textit{ART}. 
Finally, given that D3 has been annotated with argument schemes \cite{musi-etal-2018-multi}, we can explore their role in enthymeme reconstruction.
We notice that most of the generated plausible implicit premises belong to enthymemes annotated with \textit{Practical Evaluation} argument scheme, where ``the premise is an evaluation about something being ‘good’ or ‘bad’, while the claim expresses a recommendation/advice about stopping/continuing an action" (Table \ref{analysis} ).

\begin{table}[]
\centering
\small
\begin{tabular}{|p{0.4cm}|l|l|l|l|}
\hline
Data                & System                                                                & BLEU1   & BLEU2  & BS    \\ \hline
\multirow{3}{*}{D1} &  ZeroShot                                                         & 6.02  & 2.17 & 42.88 \\ \cline{2-5} 
                    &  ART                                                      & 9.16 & 3.11 & 48.35 \\ \cline{2-5} 
                    & \begin{tabular}[c]{@{}l@{}} +PARA-COMET\end{tabular} & \textbf{10.56} & \textbf{3.90} & \textbf{50.22} \\ \hline\hline
\multirow{3}{*}{D2} &  ZeroShot                                                         &   28.24    &  15.13    &  46.96     \\ \cline{2-5} 
                    &  ART                                                      &   37.77    &  18.76    &  60.63     \\ \cline{2-5} 
                    & \begin{tabular}[c]{@{}l@{}} +PARA-COMET\end{tabular} &   \textbf{44.12}    &  \textbf{24.14}    &   \textbf{67.75}    \\ \hline\hline
\multirow{3}{*}{D3} &  ZeroShot                                                         & 12.58      & 6.25     &  44.64     \\ \cline{2-5} 
                    &  ART                                                     &   14.89    &  6.34    &  51.78     \\ \cline{2-5} 
                    & \begin{tabular}[c]{@{}l@{}} +PARA-COMET\end{tabular} &   \textbf{15.56}    &  \textbf{7.50}    &   \textbf{53.38}    \\ \hline                   
\end{tabular}
\caption{\label{table:auto} Automatic evaluation of implicit premise generation by BART in 3 settings based on BLEU1, BLEU2 and BertScore(BS). Difference is significant, $( \alpha < 0.005)$ via Wilcoxon signed-rank test.}
\end{table}

\begin{table}[]
\centering
\small
\begin{tabular}{|l|l|c|}
\hline
Data                & System                                                                & Plausibility  \\ \hline
\multirow{2}{*}{D1} & ART                                                      & 50\%   \\ \cline{2-3} 
                    & \begin{tabular}[c]{@{}l@{}} +PARA-COMET\end{tabular} &   \textbf{58}\%  \\ \hline\hline
\multirow{2}{*}{D2} & ART                                                      &  48\%   \\ \cline{2-3} 
                    & \begin{tabular}[c]{@{}l@{}} +PARA-COMET\end{tabular} &  \textbf{66}\%   \\ \hline\hline                    
\multirow{2}{*}{D3} & ART                                                   &  38\%   \\ \cline{2-3} 
                    & \begin{tabular}[c]{@{}l@{}} +PARA-COMET\end{tabular} & \textbf{56}\%    \\ \hline
\end{tabular}
\caption{\label{table:human}Human evaluation results our finetuned BART models in two settings. }
\vspace{-3ex}
\end{table}

\begin{table}[h!]
\centering
\small
\begin{tabular}{|p{0.18cm}|p{1.54cm}|p{4.8cm}|}
\hline
                    &            &                                                                                                           \\ \hline
\multirow{6}{*}{D1} & St Premise & \begin{tabular}[c]{@{}l@{}}Deaf students need more specialized \\ education\end{tabular}                  \\ \cline{2-3} 
                    & St Claim   & States need special schools for the deaf                                                                  \\ \cline{2-3} 
                    & Gold       & \begin{tabular}[c]{@{}l@{}}Their parents can't always enroll them\\ in a deaf private school\end{tabular} \\ \cline{2-3} 
                    & Zero-shot   & We can't afford it, we shouldn't                                                                          \\ \cline{2-3} 
                    & ART        & \begin{tabular}[c]{@{}l@{}}The deaf students are not getting \\ enough education.\end{tabular}            \\ \cline{2-3} 
                    & +PARA-COMET     & \begin{tabular}[c]{@{}l@{}}Deaf students are not \textbf{\color{ForestGreen}being served}\\ \textbf{\color{ForestGreen}well in the schools}\end{tabular}          \\ \hline\hline
\multirow{6}{*}{D1} & St Premise & \begin{tabular}[c]{@{}l@{}}Understanding other culture is \\more important now than ever before.\end{tabular}                  \\ \cline{2-3} 
                    & St Claim   & Colleges need humanities programs                                                                 \\ \cline{2-3} 
                    & Gold       & \begin{tabular}[c]{@{}l@{}}More people now fail to understand \\other cultures\end{tabular} \\ \cline{2-3} 
                    & Zero-shot   & It's the humanities, we need them                                                                          \\ \cline{2-3} 
                    & ART        & \begin{tabular}[c]{@{}l@{}}The humanities are the most important \\subjects in college.\end{tabular}            \\ \cline{2-3} 
                    & +PARA-COMET     & \begin{tabular}[c]{@{}l@{}}There is a
 \textbf{\color{ForestGreen}lot of misinformation}\\ out there about other cultures \end{tabular}          \\ \hline\hline           
\multirow{6}{*}{D2} & St Premise & \begin{tabular}[c]{@{}l@{}}Bush new spending in 8 years? \$5.07\\ TRILLION Obama total New Spending\\ (projected out for the next 8 years)?\\\$1.44 TRILLION. And of that total,\\ only \$430 billion is non-recession \\related.\end{tabular} \\ \cline{2-3} 
                    & St Claim   & Fixed the economy                                   \\ \cline{2-3} 
                    & Gold       & Obama spends less money than Bush.                                                      \\ \cline{2-3} 
                    & Zero-shot   & We are talking about the economy                                                                     \\ \cline{2-3} 
                    & ART        & \begin{tabular}[c]{@{}l@{}}The Obama administration has spent\\  \$1 trillion.\end{tabular}                                                                                                                                                   \\ \cline{2-3} 
                    & +PARA-COMET & \begin{tabular}[c]{@{}l@{}}The Obama's \textbf{\color{ForestGreen}spending is much less}\\  than Bush's.\end{tabular}                                                                                                                                                     \\ \hline\hline
\multirow{6}{*}{D3} & St Premise & \begin{tabular}[c]{@{}l@{}}The morning-after pill has a \\number of side effects.\end{tabular} \\ \cline{2-3} 
                    & St Claim   & The morning-after pill should only be prescribed after counselling by a physician or pharmacist.,                                                                                                                                                                                                                             \\ \cline{2-3} 
                    & Gold       & Physicians and pharmacists inform about side effects.                                                                                                                                                                                                            \\ \cline{2-3} 
                    & Zero-shot   & \begin{tabular}[c]{@{}l@{}}Morning-after pills are not FDA\\ approved, they should be avoided
.\end{tabular}                                                                                            \\ \cline{2-3} 
                    & ART        & \begin{tabular}[c]{@{}l@{}}The morning- after pill can \\cause depression.\end{tabular}                                                                                                                                                   \\ \cline{2-3} 
                    & +PARA-COMET & \begin{tabular}[c]{@{}l@{}}The side effects  \textbf{\color{ForestGreen}can be very serious}.\end{tabular}                                                                                                                                                     \\ \hline                    
\end{tabular}
\caption{\label{analysis}Enthymeme generation for a given stated Premise and Claim by BART in 3 settings: zero-shot; fine-tuned on \textit{ART}; and fine-tuned on \textit{ART} + PARA-COMET. Text bolded in green displays how generations are more plausible due to incorporation of discourse aware commonsense.}
\end{table}

\section{Conclusions}
We propose an end-to-end approach for a new task of \emph{automatically generating an implicit premise given an enthymeme}. We show how leveraging abductive reasoning as an auxiliary task improves over zero-shot performance of a state-of-the-art generative language model. Finally, we build a knowledge-enhanced model by encoding discourse-aware commonsense that outperforms all existing baselines in terms of automatic metrics as well as plausibility judgements from crowdworkers. Future work includes exploring other sources for commonsense knowledge, experimenting with improved decoding techniques, as well as studying the role of argument schemes in enthymemes reconstruction.

\section{Ethical Considerations}

Although we use language models trained on data collected from the Web, which have been shown to have issues with bias and abusive language \cite{sheng-etal-2019-woman, wallace-etal-2019-universal}, the inductive bias of our models should limit inadvertent negative impacts. Unlike model variants such as GPT, BART is a conditional language model, which provides more control of the generated output. Finally, we finetune our model on the ART dataset, which is built on five sentence short stories which is devoid of harmful and toxic text especially targeted at marginalized communities.

While dual-use concerns are certainly possible here, we think that open-sourcing this technology will help to facilitate understanding of arguments with more balanced and better reasoning.  The technology should be used responsibly, particularly making sure the generation is controllable by providing the stated premise, claim and any commonsense knowledge pertaining to the enthymeme in textual form. Finally, we pay the Turkers \$15/hour, complying with minimum wage standards in US.

\bibliography{anthology,custom}
\bibliographystyle{acl_natbib}

\end{document}